%% file: main.tex
\documentclass{bmvc2k}

\usepackage{booktabs}
\usepackage{graphicx}
\usepackage{caption}
\usepackage{outlines}
\usepackage{wrapfig}
\usepackage{bm}
\usepackage{hyperref}
\usepackage{multirow}

\title{A-Scan2BIM: Assistive Scan to\\ Building Information Modeling}
\addauthor{Weilian Song}{weilians@sfu.ca}{1}
\addauthor{Jieliang Luo}{rodger.luo@autodesk.com}{2}
\addauthor{Dale Zhao}{dale.zhao@autodesk.com}{2}
\addauthor{Yan Fu}{fuyan82@gmail.com}{2}
\addauthor{Chin-Yi Cheng}{cchinyi@google.com}{3,\textdagger}
\addauthor{Yasutaka Furukawa}{furukawa@sfu.ca}{1}

\addinstitution{
 Simon Fraser University
}
\addinstitution{
 Autodesk Research
}
\addinstitution{
 Google Research
}

\runninghead{Song}{A-Scan2BIM}

\def\etal{\emph{et al}\bmvaOneDot}

\newcommand{\yasu}[1]{}
\newcommand{\weil}[1]{}
\newcommand{\rodg}[1]{}

\newcommand{\secref}[1]{Section~\ref{sec:#1}}
\newcommand{\figref}[1]{Figure~\ref{fig:#1}}
\newcommand{\tabref}[1]{Table~\ref{tab:#1}}

\newcommand{\mysubsubsection}[1]{\vspace{0.1cm} \noindent {\bf #1}:}
\newcommand{\mysubsubsectiona}[1]{\vspace{0.1cm} \noindent {\bf #1}}

\newcommand\blfootnote[1]{%
  \begingroup
  \renewcommand\thefootnote{}\footnote{#1}%
  \endgroup
}

\begin{document}

\maketitle

\begin{abstract}

% The resulting model covers a wide range of object types, with detailed 3D and semantic information.

\blfootnote{\textsuperscript{\textdagger}Work done while at Autodesk AI Lab}
This paper proposes an assistive system for architects that converts a large-scale point cloud into a standardized digital representation of a building for Building Information Modeling (BIM) applications.
The process is known as Scan-to-BIM, which
requires many hours of manual work even for a single building floor by a professional architect.
Given its challenging nature, the paper focuses on helping architects on the Scan-to-BIM process, instead of replacing them.
Concretely, we propose an assistive Scan-to-BIM system that takes the raw sensor data and edit history (including the current BIM model), then auto-regressively predicts a sequence of model editing operations as APIs of a professional BIM software (i.e., Autodesk Revit).
The paper also presents the first building-scale Scan2BIM dataset that contains a sequence of model editing operations as the APIs of Autodesk Revit.
The dataset contains $89$ hours of Scan2BIM modeling processes by professional architects over $16$ scenes, spanning over $35,000\:m^2$.
We report our system's reconstruction quality with standard metrics, and we introduce a novel metric that measures how ``natural'' the order of reconstructed operations is.
A simple modification to the reconstruction module helps improve performance, and our method is far superior to two other baselines in the order metric.
We will release data, code, and models at \url{a-scan2bim.github.io}.

%making their outputs difficult to import without manual scripting.Further, architects may need to modify or remodel an area, which would be unassisted and more cumbersome than modeling from scratch.
%
%To this end, we propose a system that allows a user to interactively edit an initial reconstruction from a state-of-the-art system, by offering suggestions that are most relevant to the latest edits.

\end{abstract}

%-------------------------------------------------------------------------

\input{sections/01-introduction}
\input{sections/02-related-works}
\input{sections/03-dataset}
\input{sections/04-method}

\input{sections/05-experiment}
\input{sections/07-conclusion}

\bibliography{biblio}
\end{document}

\maketitle

The supplementary document provides (\secref{plugin}) details of our plugin that records Revit modeling sequences; (\secref{next}) architecture specification for the next wall predictor; and (\secref{tcn}) architecture/training details for TCN used in our novel ordering metric.
Please also refer to the supplementary video demonstrating our dataset and assistive modeling system in action.
%The supplementary includes a video demonstration of our assistive system.

\section{Revit recorder plugin}
\label{sec:plugin}

To record architect modeling sequence, we created a custom Revit plugin, as Revit does not natively provide this functionality.
The plugin is written in C\#, and is triggered after each addition/modification/deletion of elements through an event handler.
Upon triggering, the full state of the affected elements is accessible through the Revit API and saved.
By exploiting C\#'s Reflection functionality, we can then automatically iterate through the element object's properties and write them to a JSON file.
We also modified the auto-save functionality of Revit so we save the state of the model every 15 minutes. The JSON files are also split into 15-minute sessions as a fail-safe in case of any errors during recording.

To obtain an equivalent API call that mimics a modeling operation, we find the difference between the saved element states, and translate it programmatically using the Revit API.
A second custom Revit plugin is created for the purpose of replaying the recorded data step-by-step, which is also demonstrated in the video.

\begin{figure}
    \centering
    \includegraphics[width=0.8\textwidth]{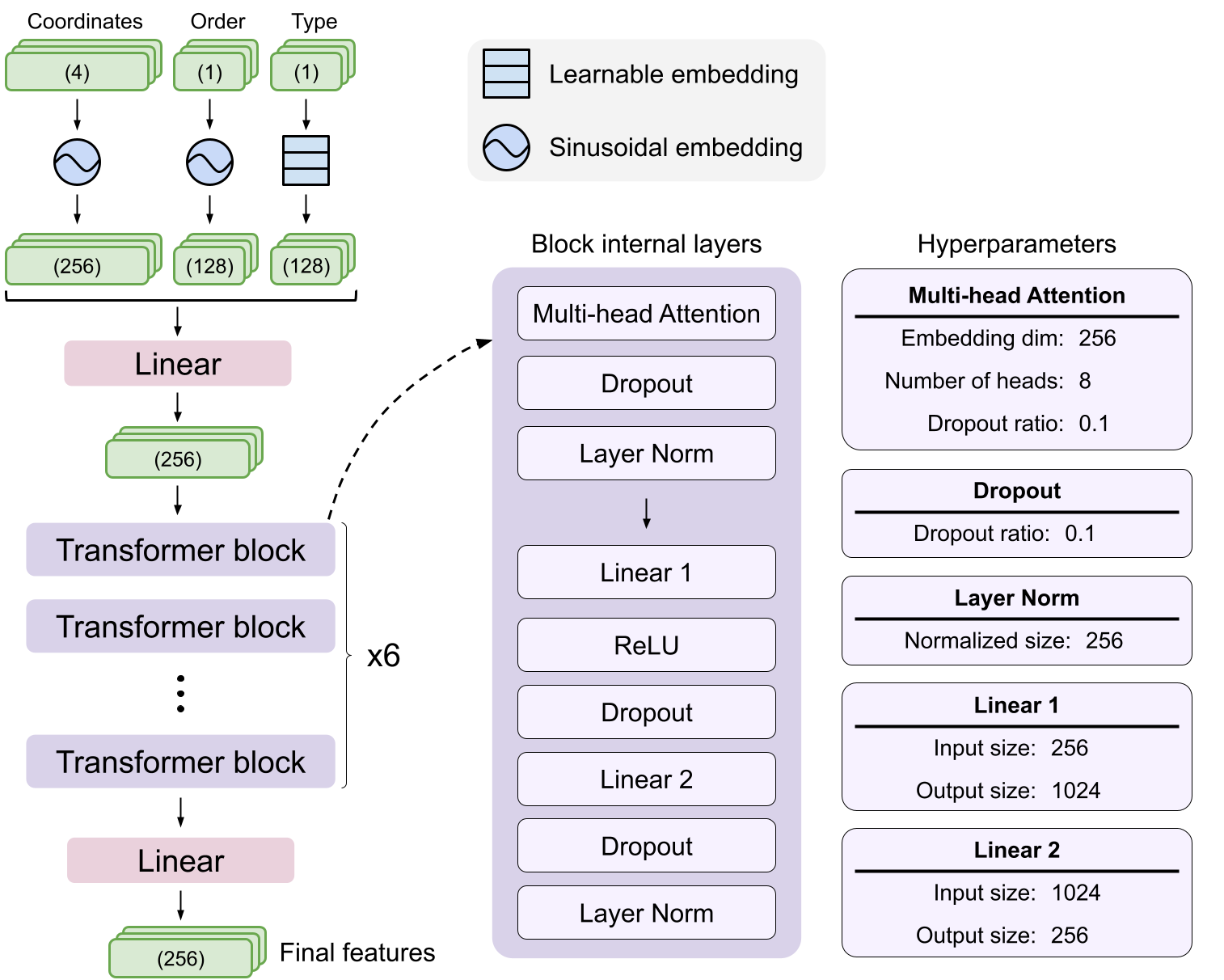}
    \vspace*{.3cm}
    \caption{
    Next wall predictor architecture. Note that our transformer block is similar to that of~\cite{chenHEATHolisticEdge2022}, except without the deformable image attention module.
    }
    \label{fig:next_wall}
\end{figure}

\section{Next wall predictor architecture}
\label{sec:next}

\figref{next_wall} shows the high-level architecture of our next wall predictor.
Each wall node contains three information: the wall endpoint coordinates $(x_0, y_0, x_1, y_1)$, its timestep $t$, and one of three wall type encodings.
The number in parenthesis in the green boxes denotes feature dimension, and the two outer linear layers map dimensions from $512\rightarrow256$ and $256\rightarrow256$ respectively.
The Transformer network is a stack of six blocks, with each block consisting of a mix of multi-head attention, dropout, layer normalization, linear, and ReLU activation layers.
The transformer block is similar to that of the one proposed in~\cite{chenHEATHolisticEdge2022}, except that we removed any modules related to deformable image attention as our prediction network relies on geometry only.
In the figure we provide the full details for ease of reproducibility.
% This includes the internal layer configurations and hyperparameters such as hidden feature dimension, number of heads in multi-head attention, dropout ratios, linear layer output sizes, etc.
% We refer to their code base~\cite{chenHEATHolisticEdgeCode2023} for more details.

\section{TCN architecture and training details}
\label{sec:tcn}

To obtain a latent encoding for an arbitrary modeling sequences, we train a Temporal Convolutional Network (TCN) to auto-encode wall coordinates and extract hidden features from a middle layer.
Specifically, we borrow the implementation from~\cite{BaiTCN2018}, and use the code for the ``Copy Memory Task''.
During training, we construct an example by sampling between 2 to 10 edges from a random floor.
We flatten the wall coordinates to a one-dimensional vector as network input and use the L2 distance as the reconstruction loss.
Batch size of 32 and Adam optimizer is used with learning rate $5e^{-4}$, and we train until convergence, around $20\,000$ iterations.
During inference, we flatten the features after the 6\textsuperscript{th} layer as the sequence encoding to be used for the order metric.

\bibliography{biblio}

%% file: sections/01-introduction.tex
\section{Introduction}
\label{sec:intro}
% -----------------

% [ Motivation ]
% Explain what exciting things happen when Scan-to-BIM becomes possible. Or tell a story of why this is important.
% - Explain the task of Scan-to-BIM and what BIM models are

Building Information Modeling (BIM) serves as a modern foundation for building design, construction, and management. This comprehensive approach involves generating a complete digital representation of a building, integrating various engineering disciplines such as architecture, electrical, HVAC, and more.
BIM transcends traditional CAD modeling by incorporating not only geometric data but also essential constraints and metadata.
%required for diverse industries like Civil, Construction, MEP (Mechanical, Electrical, and Plumbing), among others. 
The process of creating a BIM model from a 3D scan of an existing building is referred to as Scan-to-BIM.
Leveraging a BIM model significantly simplifies cost-saving assessments, such as heating optimization and structural analysis. Furthermore, it streamlines renovations by providing a consistent, underlying model for designers across all industries.
%, facilitating real-time collaboration.
%
Swiftly generating an architectural model through Scan-to-BIM serves as the initial step for diverse industries like Civil, Construction, MEP (Mechanical, Electrical, and Plumbing), among others.

Scan-to-BIM is a labor-intensive process that often demands numerous hours of manual work from a professional architect, even for a single building floor. There exists commercial and open-source software for automatic Scan-to-BIM, but in general architects do not use these software due to poor performance and integration with their workflows.
Recognizing this challenge, the paper aims to assist architects in the Scan-to-BIM process rather than replace them entirely. Nonetheless, existing building reconstruction algorithms struggle with this assistive task, as their reconstruction approach differs significantly from that of an architect. Specifically, architects create a BIM model by executing a series of modeling editing operations using CAD software, while current algorithms~\cite{liuRastertoVectorRevisitingFloorplan2017, chenFloorSPInverseCAD2019, zhangConvMPNConvolutionalMessage2020, chenHEATHolisticEdge2022} typically reconstruct a model in a single step or sequentially but without enabling human-interactions.
%This task is normally performed by a professional architect, using a BIM software to directly model with the point cloud overlaid. The modeling process is highly precise, and can take upwards of 10 hours per floor. 
%In the vision community, there has been remarkable advancements in structured reconstruction algorithms, although even the state-of-the-art system does not meet the required precision needed.
%Architects do not trust outputs from any one-shot reconstruction system, and it is often easier to model from scratch instead.

This paper proposes an assistive Scan-to-BIM system that takes raw sensor data and edit history (including the current BIM model), then auto-regressively predicts a sequence of model editing operations as APIs for professional BIM software, specifically Autodesk Revit.
The sequence is presented to the user in the Revit interface, who can either accept or reject the suggestions for other options.
%
%Scan-to-BIM requires many hours of manual work, even for a single building floor by a professional architect. %Given its challenging nature, the paper focuses on helping architects on the Scan-to-BIM process, instead of replacing them completely. However, existing building reconstruction algorithms fail on the task, because their process of reconstruction is far from that of an architect. Concretely, an architect makes a BIM model by performing a sequence of modeling editing operations by using a CAD software. On the other hand, existing algorithms usually reconstruct a model in one shot. Concretely, we propose a semi-automatic Scan-to-BIM system that takes the raw sensor data and historical edits (including the current BIM model), then auto-regressively predicts a sequence of model editing operations as APIs of a professional BIM software, in particular, Autodesk Revit.
We focus on wall reconstruction, one of the main steps of Scan-to-BIM workflow taking up $80\%$ of the modeling steps based on our data collection process with architects.
%
%Specifically, given a 3D point cloud of one floor, we retrieve the centerlines and the thickness of walls.
%The point clouds are of centimeter-level accuracy, and covers floors thousands of square meters in size.
% [ Our research ]
% We should briefly summarize our research that directly addresses the challenges above.
%Our research instead proposes a different solution: SEABIM, a novel semi-automatic system for wall reconstruction.
%The SEABIM is designed to speed up manual modeling by considering all past editing steps and offering accurate, relevant wall suggestions multiple steps into the future.
Concretely, the system operates in two stages.
%It operates in two stages.
First, we use a modified version of a state-of-the-art floorplan reconstruction system~\cite{chenHEATHolisticEdge2022} to enumerate candidate walls. The modification is to estimate wall thickness and scale to building-scale scans.
%candidate wall predictions are obtained using a modified version of a state-of-the-art floorplan reconstruction system HEAT~\cite{chenHEATHolisticEdge2022} to account for wall thickness and the large-scale nature of our data.
%
Secondly, an auto-regressive transformer network processes a set of candidate walls and the addition action history (with order information) to predict a future sequence of actions.
The transformer learns a feature embedding for a candidate wall with a contrastive loss, such that the next action is closer to the latest one in the feature space.
% \yasu{write a condensed sentence giving the technical core}.
%
The wall addition action activates a corresponding wall addition API via a Revit Plugin, where nearby wall segments are automatically joined and elevated to 3D. These segments can also be interactively modified within Revit. We have collected the first building-scale Scan2BIM dataset, comprising 89 hours of modeling processes by architects across 16 scenes over 35,000 square meters.

%The auto-regressive network learns from human modeling sequences we collected from architects, using a custom plugin developed for Revit, the most popular BIM software by far.
%The network ensures that only relevant walls are suggested, which is beneficial when either modeling from scratch or from an initial reconstruction.
%

% [ Summary of results and potentially summary of contributions ]

We have included a supplementary video showcasing our system assisting a user in a real modeling scenario.
We also conduct qualitative and quantitative evaluations of the proposed system against several baselines using 1-fold cross-validation. In addition to standard metrics for structured reconstruction~\cite{chenHEATHolisticEdge2022}, the paper assesses the "naturalness" of the reconstruction order by introducing a new metric inspired by the FID score, a standard metric for generative models~\cite{nauata2020house}. Our experiments demonstrate that the proposed system more closely resembles approaches by architects than the baselines. In summary, the paper's contributions are threefold: 1) A transformer network with contrastive loss training that predicts a natural sequence of actions; 2) A Scan2BIM assistive system that directly drives professional CAD software throughout the BIM reconstruction process; and 3) A building-scale Scan2BIM dataset containing 89 hours of BIM modeling sequences by professional architects. To further promote the development Scan-to-BIM techniques, we will release all the data, models, and code.

%% file: sections/02-related-works.tex
\section{Related works}
\label{sec:related}

% [ Structured reconstruction and invese cad, where our latest papers should give good references. But do not just copy related work section of our previous papers. You have to write with your own words. ]

Our paper introduces a new dataset of large-scale buildings, and proposes an interactive algorithm for structured reconstruction.
We review datasets and methods related to architectural structured reconstruction, along with the literature of Scan-to-BIM.

\mysubsubsectiona{Architectural structured reconstruction datasets}
have been introduced for modeling at the scene-level, floor-level, and city-level. Scene-level primarily focuses on semantic segmentation~\cite{armeni3DSemanticParsing2016, songSemanticSceneCompletion2017, changMatterport3DLearningRGBD2017, daiScanNetRichlyAnnotated3D2017}, plane reconstruction~\cite{changMatterport3DLearningRGBD2017, daiScanNetRichlyAnnotated3D2017}, and wire-frame parsing~\cite{huangLearningParseWireframes2018}, while floor-level~\cite{ikehataStructuredIndoorModeling2015, liuRastertoVectorRevisitingFloorplan2017, kalervoCubiCasa5KDatasetImproved2019, nauataVectorizingWorldBuildings2020} aims to obtain architectural structures such as walls and columns.
HoliCity is a city-level dataset containing 3D structures of buildings along with street-level panoramic images and segmentation masks.
In terms of data type, the floorplan dataset used by HEAT~\cite{chenHEATHolisticEdge2022} is closely related to our dataset, as both the input is a point cloud and the output is a planar graph of edges. However, our floors are significantly larger and have more complex structures with walls of varying thickness. In terms of floor scale, S3D proposed by Armeni~\etal~\cite{armeni3DSemanticParsing2016} more closely resembles our dataset, as both consist of large office spaces. Our labels contain significantly more detailed geometry, as our dataset represents architectural models. Additionally, our dataset includes 16 floors compared to 5, covering over $35\,000$ square meters in comparison to $6,000$ square meters in their dataset.

%\mysubsubsectiona{Architectural structured reconstruction datasets} have been introduced for modeling at the scene-level and floor-level. The former primarily deals with semantic segmentation~\cite{armeni3DSemanticParsing2016, songSemanticSceneCompletion2017, changMatterport3DLearningRGBD2017, daiScanNetRichlyAnnotated3D2017}, plane reconstruction~\cite{changMatterport3DLearningRGBD2017, daiScanNetRichlyAnnotated3D2017} and wire-frame parsing~\cite{huangLearningParseWireframes2018}; while datasets~\cite{ikehataStructuredIndoorModeling2015, liuRastertoVectorRevisitingFloorplan2017, kalervoCubiCasa5KDatasetImproved2019, nauataVectorizingWorldBuildings2020} proposed for the latter task is concerned with obtaining architectural structures such as walls and columns. In terms of data type, the floorplan dataset used by HEAT~\cite{chenHEATHolisticEdge2022} closely matches with our dataset, as the input is a point cloud and output is a planar graph of edges. However, our floors are magnitudes larger, and are more complex in structure with walls of different thickness. In terms of floor scale, S3D proposed by Armeni~\etal~\cite{armeni3DSemanticParsing2016} more closely resembles our dataset, as both are of large office spaces. Our labels contains significantly more detailed geometry in comparison, as ours is an architectural model. Besides, our dataset contains 15 floors compared to 5, covering over ??? square meters compared to $6000$. \weil{Some more numbers to confirm here}

\mysubsubsectiona{Architectural structured reconstruction methods}
commonly use input data such as images, RGB-D scans, or 3D point clouds, and output the man-made structure in a planar graph representation. A two-stage pipeline has been the dominant approach to recover building structures, where geometry primitives such as corners are first detected, followed by their assembly process.
Ikehata~\etal~\cite{ikehataStructuredIndoorModeling2015}, Liu~\etal~\cite{liuRastertoVectorRevisitingFloorplan2017}, and Chen~\etal~\cite{chenFloorSPInverseCAD2019} employed optimization systems for the second stage, using predefined grammar, integer programming, or energy minimization techniques. These methods typically make strong assumptions about the structure, such as a Manhattan layout. Conv-MPN~\cite{zhangConvMPNConvolutionalMessage2020} improved upon previous works by proposing a fully-neural architecture, allowing for learnable topology inference. MonteFloor~\cite{stekovicMonteFloorExtendingMCTS2021} incorporated Monte Carlo Tree Search while remaining fully differentiable and removing assumptions in earlier optimization-based methods. Finally, HEAT~\cite{chenHEATHolisticEdge2022} achieved state-of-the-art outdoor and indoor reconstruction results through a transformer architecture~\cite{vaswaniAttentionAllYou2017}, which learns to classify edge proposals by considering edge image features and global topology.
3D wireframe techniques~\cite{ xuLineSegmentDetection2021, zhouEndtoEndWireframeParsing2019, zhouLearningReconstruct3D2019, huangLearningParseWireframes2018} often do not rely on primitives and directly predict the target geometries from learnable embeddings.

\mysubsubsectiona{Scan-to-BIM}
% To our knowledge, there are papers in computer vision that claim ``Scan to BIM'', but they merely reconstruct a standard structured geometry, just like the techniques summarized in the previous section. In the civil engineering community, there exist real Scan-to-BIM systems but they do not focus on technical contributions.
There exists automatic Scan-to-BIM systems in the Civil Engineering community~\cite{jungAutomated3DVolumetric2018, ochmannAutomaticReconstructionFully2019, nikoohematIndoor3DReconstruction2020}.
They often employ multi-step optimization systems such as semantic segmentation followed by plane fitting and wall reconstruction.
To our knowledge, they are not evaluated on standard datasets and no code exists for comparison.
Exceptions are with works from Bassier~\etal~\cite{bassierTopologyReconstructionBIM2020, bassierUnsupervisedReconstructionBuilding2020}, where automatic systems are proposed and evaluated on the S3DIS~\cite{armeni3DSemanticParsing2016}.
Furthermore, these papers focus on system integration instead of technical contribution, which would be more critical for the computer vision community.

%% file: sections/03-dataset.tex
\section{Assistive Scan-to-BIM dataset}
\label{sec:dataset}

\begin{figure}
    \centering
    \includegraphics[width=\textwidth]{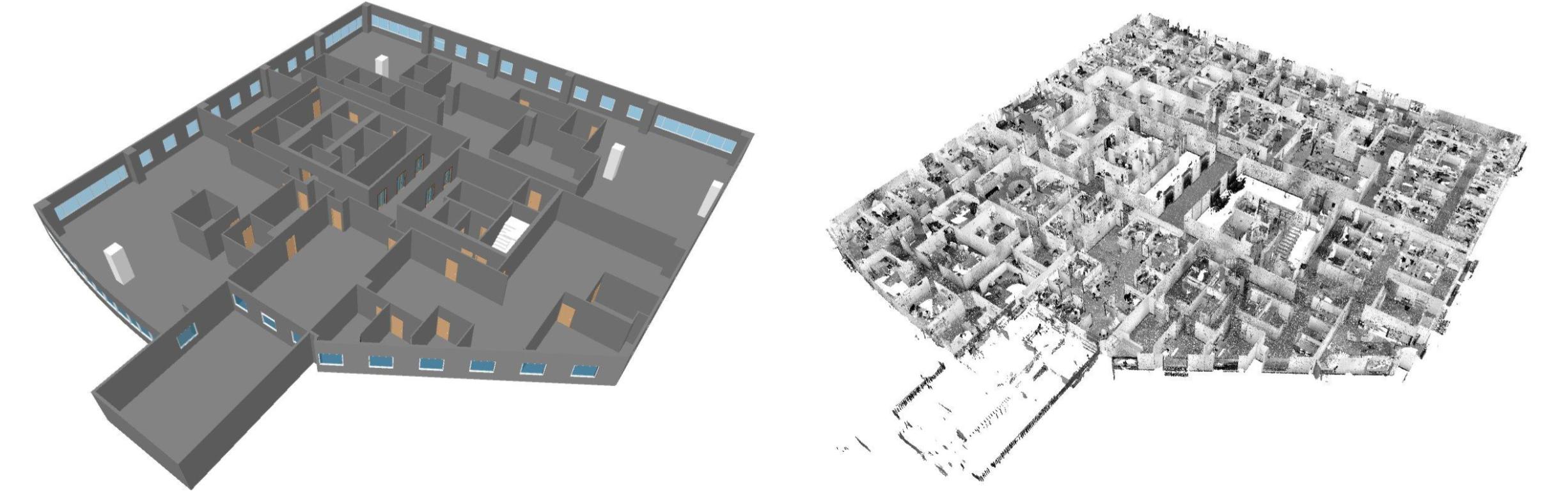}
    \vspace*{.05cm}
    \caption{
    A BIM model and a scan from one floor with the ceiling removed.
    Walls, doors, windows, and columns account for $84.2\%$, $1.1\%$, $14\%$, and $0.6\%$ of the modeling steps.
    }
    \label{fig:dataset}
\end{figure}

\begin{table}[]
\centering
\caption{Statistics of our proposed dataset. Average refers to average per floor.}
\label{tab:stats}
\vspace*{.3cm}
\resizebox{\textwidth}{!}{%
\begin{tabular}{lccllccllcc}
\multicolumn{3}{c}{General stats} &  & \multicolumn{3}{c}{Element counts} &  & \multicolumn{3}{c}{Element type counts} \\ \cline{1-3} \cline{5-7} \cline{9-11} 
                      & Total & Average &  &         & Total & Average &  &         & Total & Average \\ \hline
Floor size ($m^2$)    & 35528 & 2221    &  & Walls   & 3142  & 196     &  & Walls   & 189   & 12      \\
Annotation time (hrs) & 89    & 6       &  & Doors   & 808   & 51      &  & Doors   & 297   & 19      \\
Modeling steps        & 45306 & 2832    &  & Windows & 3344  & 209     &  & Windows & 77    & 5       \\
JSON data size (GB)   & 29.97 & 1.87    &  & Columns & 323   & 20      &  & Columns & 113   & 7       \\ \hline
\end{tabular}%
}
\end{table}

% \usepackage{booktabs}
% \usepackage{graphicx}

% What is BIM?
% What are our inputs?
% What is our target data to collect?
% Describe what is BIM, what are the point cloud inputs, and our goal of obtaining edit sequence.

% Statistics to report:
% Number of floors
% Point cloud resolution
% Total/average size of floors
% Total/average of annotation time
% Total/average of modeling steps
% Total/average number of elements
% Number of element types
% Percentage of steps w.r.t. element types

We borrow building-scale scans from the ``Computer Vision in the Built Environment'' workshop series at CVPR~\cite{cv4aec}. We have used 16 scans from 11 buildings, with space types including office spaces, parking lots, medical offices, and laboratories.
Point clouds are captured using professional surveying equipment, which is accurate to 1 cm. To obtain ground-truth BIM models, we hired 5 professional architects and asked them to model at Level of Development (LOD) 200, which includes floors, walls, doors, windows, columns, and stairs. Autodesk Revit software was used for modeling, which will be our target platform.

% \rodg{the following paragraph about the Revit plugin looks a little bit too detailed for me. Maybe move some of the details to the appendix?}
The main goal is to record modeling sequences, while
%architect modeling operations.
Revit does not provide this functionality. We created a custom Revit plugin that saves the states of the BIM model before and after an operation, and programmatically translates the state difference into an equivalent operation as a Revit API call.
%To achieve this, we created a custom Revit plugin for recording, as Revit does not natively provide this functionality. The plugin is written in C\#, and is triggered after each addition/modification/deletion of elements through an event handler. Upon triggering, the full state of the affected elements is accessible through the Revit API and saved.
%
% By exploiting C\#'s Reflection functionality, we can then automatically iterate through the element object's properties and write them to a JSON file.
%To obtain an equivalent API call that mimics the modeling operation on an element, we find the difference between element states, and translate it programmatically.
% A second custom Revit plugin is created for the purpose of replaying the recorded data step-by-step.
Please see the supplementary for more details.

\tabref{stats} shows the statistics of our dataset. \figref{dataset} visualizes a BIM model and a scan.
% We highlight that the percentage of 
Walls, doors, windows, and columns account for $84.2\%$, $1.1\%$, $14\%$, and $0.6\%$ of the operations, respectively.
% showing that wall modeling takes up a majority of the steps.
A 3-channel top-down point density image is an input to our system: We slice the point clouds horizontally at 6.56, 8.2, and 12 feet above the ground, and calculate the point density within the three slices in the top-down view at one inch resolution.

%% file: sections/04-method.tex
%\section{Semi-automatic modeling pipeline}
\section{Assistive Scan-to-BIM}
\label{sec:method}

\begin{figure}
    \centering
    \includegraphics[width=\textwidth]{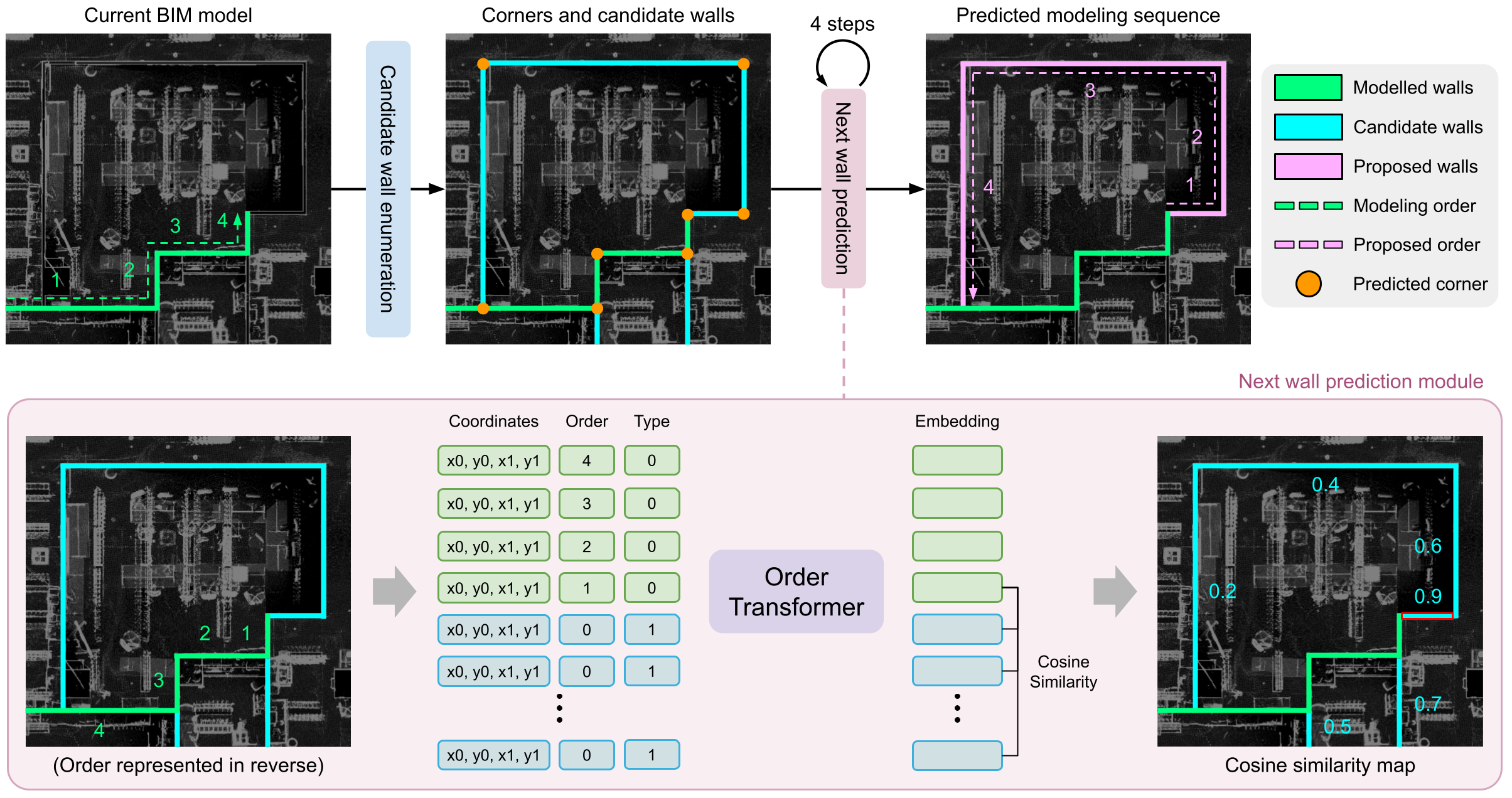}
    \vspace*{.05cm}
    \caption{
    System overview. Given an existing BIM model as the current state, we first obtain wall candidates by enumerating corners and edges with thickness. We then auto-regressively predict an ordering to the candidate walls using a Transformer network.
    }
    \label{fig:method}
\end{figure}

Scan-to-BIM turns sensor data into a BIM model, where we focus on wall structures. As a BIM assistant, the proposed system is integrated with a BIM software, in our case Autodesk Revit. The section explains 1) The pre-processing module, in which we enumerate wall segment candidates by modifying the current state-of-the-art floorplan reconstruction system~\cite{chenHEATHolisticEdge2022} to handle building-scale scans; 2) The next wall prediction module, which is the core of our system; and 3) The assistive Scan-to-BIM system that integrates the two modules with Revit.
We refer to \figref{method} for an overview of our system.

%to handle building scale scans instead of a single house;

%, and our system takes sensor data and auto-regressively reconstructs a sequence of walls, each of which is added to the BIM model via a Revit API. 

%%%%%%%%%%%%%%%%%%%%%%%%%%%%%%%%%%%%%%%%%%%%%%%%%%%%%%%%%%%%%%%%%%%%%%%%%%%%%%%%%%
\subsection{Candidate wall enumeration} \label{sec:candidate_wall}

Given a density image and existing walls, we use the state-of-the-art floorplan reconstruction system HEAT~\cite{chenHEATHolisticEdge2022} with a few key modifications to enumerate wall candidates in four steps:
%the module enumerates wall candidates
%candidate wall locations and thickness. The module operates 
%in four steps: 
corner enumeration, wall enumeration, wall thickness prediction, and duplicate removal.
% As are modifying from Chen~\etal's system~\cite{chenHEATHolisticEdge2022}, we refer to their paper for details regarding hyperparameter choices.

\mysubsubsectiona{Corner enumeration} is done by a HEAT corner detection module that uses a Transformer network to estimate the corner likelihood at every pixel and apply non-maximum suppression (NMS). Since our density image is large, we divide the image into $256\times 256$ local windows with $64$ pixels overlap, compute the pixel-wise likelihood by the HEAT module, and merge results while keeping the maximum value at the overlaps. The same NMS filter applies.

%are obtained using a Transformer network, where each input node is a pixel location of the input image and output is the corner likelihood.
%The pixel location is a 128-dimensional sinusoidal encoding of the $(x,y)$ coordinate. Self-attention between pixels is disabled, and cross-attention with image features from a CNN is achieved through deformable attention.
% Due to memory limitations with the large-scale point density image, we train the corner network in local crops and then merge the predictions.
% We divide the large density image into $256\times256$ crops, with $64$ pixels of overlap in between.
% After inference on each crop, the resulting heatmaps are pasted onto a global canvas by finding the maximum between the two.
%Non-max Suppression is then applied to obtain predicted corners.

\mysubsubsectiona{Wall enumeration} is done by a HEAT edge classification module, 
%
%first exhaustively enumerates all pairs of predicted corners to obtain edge candidates. The edge candidates are encoded into 256-dimensional sinusoidal embeddings as input to two weight-sharing Transformer networks. The networks perform self-attention among edge geometry features, and only one network (branch) is allowed to attend to image features in order to enforce geometric reasoning.
where we modify the deformable attention layer to pool image features along very long edges in our task.
%to better attend to image features of very long edges.
Concretely, instead of sampling image features from one reference point (typically the edge center), we sample from $N$ linear-interpolated reference points along an edge, and apply max-pooling.
%perform max-pooling for the final feature.
%
%Finally, we perform binary classification on the final edge features through a two-layer MLP to obtain predicted edges.
%
Lastly, Revit represents a T-junction as one long edge and one short edge. During training, we split the long edge into two at the junction point to obtain a proper planar graph structure.
%During training, we split long edges in our dataset at junction points to obtain the proper planar graph representation.

\mysubsubsectiona{Wall thickness prediction}
is obtained by an extra two-layer MLP which is added to the end of the image branch of the edge classifier above. The thickness can range from 1 inch to 84 inches with an increment of an inch.
%We classify the thickness to be from 1 inch  to 48 inches in an increment of an inch.
During training, we apply cross-entropy loss on all walls with known thickness while ignoring the rest.

\mysubsubsectiona{Duplicate removal}
looks through the enumerated walls and retains only ones that do not exist yet.
Given the coordinates of a pair of enumerated and existing walls, we compute the distances of their end-points.
%from one edge's two corners to the other edge.
An enumerated wall is flagged as a duplicate and removed, if both corners are less than 10 pixels for at least one existing wall. We also perform the same filter between the enumerated walls themselves and remove duplicates.
%If the distances from both corners are less than 10 pixels, we remove the predicted wall.
%We also perform the same filter between the predicted walls themselves, to remove any self-overlaps.
%The resulting set of walls are the candidates for the next wall prediction module.

\subsection{Next wall prediction}
\label{sec:infer_seq}
% Given modelled walls with ordering and candidate walls as input, this module predicts a feature embedding for each wall, such that the distance between the current and a candidate wall indicates the likelihood of being the next step.
Given an existing set of walls optionally with their edit history (i.e., the sequence in which they were added to the model) along with a set of wall candidates, we calculate a score for each candidate. The score estimates the likelihood of each candidate wall being the next addition to the model. The process runs auto-regressively to obtain future modeling sequences of arbitrary length.
Please see the lower half of~\figref{method} for a high-level overview.

A Transformer network is our architecture where an existing or a candidate wall is a node. 
The network contains six blocks of self-attention layers, and produces a 256-dimensional embedding vector for each node.
The cosine similarity in the embedding space between the last existing wall to candidate walls determines the scores of the next one to be added (i.e., higher the more likely).
We refer to the supplementary for the full architecture specifications.

Each wall node is a concatenation of three embeddings:
1) 256-dimensional sinusoidal embedding of the wall coordinates;
2) one of three learnable 128-dimensional wall type embeddings; and
3) 128-dimensional sinusoidal embedding of the timestep $t$ when the wall was added.
The timestamp ($t$) is assigned in reverse chronological order. Specifically, candidate walls are marked with $t=0$. The last added wall is assigned $t=1$. The second last added wall is given $t=2$, and so on. For walls that have not been modified in the last 10 steps, we designate their timestamp as $t=10$.
%The timestamp ($t$) is set in a reverse order. That is, candidate walls have  $t=0$. The last modified wall has $t=1$. The second to last has $t=2$, etc.
%For existing walls older than $10$ steps, we set $t=10$.
%For $t$, we encode in reverse order such that candidate walls have $t=0$, the last modified wall has $t=1$, second to last has $t=2$, etc.
%
%We consider all walls modelled over $10$ steps ago to have $t=10$, as fine-grained order past $t\geq10$ is not helpful.
%
For type embeddings, the three types of walls are ones with timestep $t=0$, $1\leq t < 10$, and $t\geq10$, respectively.
The concatenated embedding is then projected down to the dimension of 256 by a linear layer.

We train the Transformer network with a contrastive loss by constructing triplets with the last wall ($t=1$), the ground-truth (GT) next wall, and all other candidate walls, such that the cosine embedding distance between the latest and GT next wall are closer than the rest of the candidates by a margin of 1.
We train on GT edges and sequences, and found that the network also works well on our reconstruction results during test time.

\subsection{Assistive Scan-to-BIM}
% \yasu{Explain maybe fully automatic version first.}
% \yasu{Then, how to run a system when there are user interactions}

The final assistive system combines the above modules and works with or without user interaction, from scratch or an existing BIM model.
%Finally, we explain the full assistive system, which can be used with or without user interaction, and either modelling from scratch or from an existing BIM model.
Regardless of the scenario, we first run the corner enumeration module and cache the result to be used in all subsequent steps.
%enumerate corners 
%we No matter the scenario, we cache corner predictions once at the beginning of modeling, and use them for all subsequent steps.

% Our system takes the sensor data and the BIM model under construction, then predicts the sequence of future wall addition actions auto-regressively. 

% The proposed system is designed to work with a CAD system, in our case, Autodesk Revit, where a building model is stored inside Revit. The wall structure is represented as a set of wall segments and the order of being added to the model, where the order information may not be available for some or all the segments when importing an existing BIM model.

\mysubsubsectiona{Automatic mode} reconstructs the BIM model without user-interaction by auto-regressively running the next wall prediction network, while taking the candidate wall with the highest score every time.
%The user may choose to reconstruct in one-shot by simply running the wall enumeration module and add all candidate walls.
%Optionally the user can ask the system to reconstruct walls step-by-step and pause anytime to apply manual correction. For step-by-step reconstruction, the system randomly chooses one wall as the last modified wall and run the next wall prediction module auto-regressively.
Note that when starting from an existing model, we do not have the edit history information and set $t=10$ for all existing walls.
%where we don't have timestep information, we set $t=10$ for all existing walls.

\mysubsubsectiona{Assistive mode} aims to speed up manual modeling by offering wall auto-completions after each user interaction.
A user interaction can be either a corner addition or wall addition/modification action, the former of which is a new action implemented by our plugin.
Each interaction introduces new or modified corners, and we combine them with our cached corners and re-run the wall enumeration module to update the wall candidates.
%as input to the wall enumeration module.
%
The system then auto-regressively predicts a future modeling sequence of length-$N$ ($N$ changeable by a user), and displays the sequence as special lines in Revit.
The user can choose to accept the proposal, or instead change the modeling direction by requesting the top-3 next wall predictions and choosing one as the next step.
Auto-completion then resumes. Please see the supplementary video for the demonstration of the mode.

%% file: sections/05-experiment.tex
\section{Experiments}
\label{sec:experiments}

\begin{figure}
    \centering
    \includegraphics[width=\textwidth]{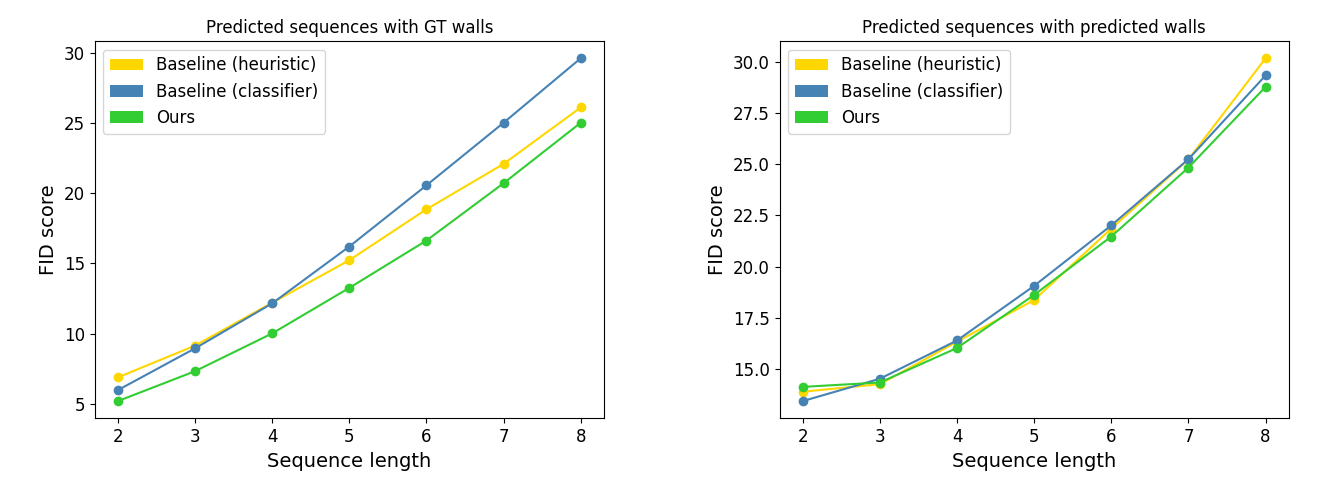}
    \vspace*{.05cm}
    \caption{
    Main results, evaluating modeling sequence generated by different methods at different sequence lengths. Lower is better.
    }
    \label{fig:seq_vs_fid}
\end{figure}

We developed our system in Python 3.8 and PyTorch 1.12.1. Training utilizes a single NVIDIA A100 GPU with 40GB of memory.
For corner and edge network hyper-parameters, we refer to the HEAT model~\cite{chenHEATHolisticEdge2022}.
The learning rate for all networks is $2\times10^{-4}$. The batch sizes are 8, 1, and 128 for the corner, edge, and next wall prediction networks, respectively. For the edge network, we perform gradient accumulation every 16 steps, effectively yielding a batch size of 16.
The training takes $93,000$, $210,000$, and $200,000$ steps for the three networks, with learning rate decay by a factor of $0.1$ every $53,000$, $70,000$, and $100,000$ steps, respectively.
For data augmentation, we use random rotation for corner training. For wall enumeration, we normalize the image and edges so that the longest edge does not exceed 1000 pixels, after which we apply random rotation and scaling. For the next wall prediction, we first center and normalize the edges by a maximum length of 1000, then perform random translation, rotation, and scaling.
After fine-tuning the corner detector of HEAT, we obtain corner precision/recall of $83.70$/$71.10\%$ under matching threshold of 30 inches.

%Our implementation is written with Python 3.8 and PyTorch 1.12.1. We use one NVIDIA A100 GPU with 40GB of memory for training. We refer to HEAT~\cite{chenHEATHolisticEdge2022} for hyper-parameters of the corner and edge networks. Learning rate for all networks is $2e^{-4}$. We employ batch size of 8, 1, and 128 for the corner, edge, and next wall prediction network. For the edge network, we perform gradient accumulation every 16 steps, for an effective batch size of 16. We train the networks for $93\,000$, $210\,000$, and $200\,000$ steps, and decay the learning rate by $0.1$ every $53\,000$, $70\,000$, and $100\,000$ steps respectively. For corner training, we perform augmentation by random rotation. For edge training, we normalize the image and edges such that the longest possible edge is no more than 1000 pixels long, and then perform random rotation and scaling. For next wall training, we center and normalize the edges by maximum length of 1000, followed by random translation, rotation, and scaling.

\subsection{Baseline methods}

\mysubsubsection{Heuristic} 
% A greedy algorithm can be used to determining the sequence of wall additions, where the candidate wall nearest to the last edit is selected.
To determine wall addition sequence, one can greedily pick the candidate wall nearest to the last edit.
Specifically, for each pair of walls, we identify the closest points and calculate their distance.
% We apply this process from the last wall in the sequence to all candidate walls. The candidate wall exhibiting the smallest distance is then selected as the next step in the sequence.
We apply this process from the last wall in the sequence to all candidate walls, choosing the candidate wall with the smallest distance.
If there is a tie, we randomly choose one wall among the closest candidates.
%As a naive solution to obtaining order, one can greedily pick the candidate wall that is closest to the last edit. Specifically, given a pair of walls, we find the nearest points between them and compute their distance. We do so from the last wall in the sequence to all candidate walls, and choose the one with the smallest distance as the next step.

\mysubsubsection{Classifier} 
In a more straightforward approach, one could train a classifier to distinguish between valid and invalid sequences. This classifier could enumerate all potential sequence candidates, selecting the one with the highest probability. 
To train such a classifier, we construct positive examples by considering all sub-sequences of the ground truth modeling steps, starting from the first step.
For every positive sequence, we generate a negative example by substituting the last wall with one from a future step.
To overcome class imbalance issue, we replicate positive examples to that of the negative ones. 
In terms of binary classification, we utilize the same architectural framework as the next wall prediction module. However, we introduce an extra classification token into the transformer network. This token undertakes self-attention with the walls, and a final linear layer is used to output the binary probability.
\input{tables/attn_samples}
\input{tables/entropy_acc}

\subsection{Novel order metric}
Evaluating next wall prediction solely based on accuracy is overly stringent, given the ambiguous nature of the task. We propose a new metric to assess the "naturalness" of the predicted wall modeling sequence. This metric draws inspiration from the Fréchet Inception Distance (FID) score \cite{heuselGANsTrainedTwo2017}, a measure commonly used to evaluate the quality of samples produced by image generation models.
Given two sets of real and predicted wall sequences, we initially generate a latent encoding for each sequence using a Temporal Convolutional Network (TCN)~\cite{BaiTCN2018}. The TCN is trained to auto-encode random subsets of ground truth edges, with the output from the 6\textsuperscript{th} hidden layer being used.
For each set of encodings, we calculate two Gaussian distributions that capture the mean and variance of each latent dimension. The final score is determined by computing the Fréchet distance between these two distributions. A lower score indicates a better model.
For more detailed information regarding the TCN architecture and its training process, please refer to the supplementary materials.
%Evaluating next wall prediction in terms of accuracy is a harsh metric, as it is often an ambiguous task. We therefore propose a new metric to evaluate how ``natural'' the predicted wall modeling sequence is. Our metric is inspired by the FID score~\cite{heuselGANsTrainedTwo2017}, commonly used to evaluate samples from image generation models. Given two sets of real and predicted wall sequences, we first obtain a latent encoding of each sequence with a Temporal Convolutional Network (TCN)~\cite{BaiTCN2018}. The TCN is trained to auto-encode random subsets of GT edges, and we use the output after the 6\textsuperscript{th} hidden layer. Two Gaussian distributions are computed for each set of encodings, capturing the mean and variance of each latent dimension. Finally, the Fr\'echet distance is computed between the two distributions to obtain the final score, where lower is better. Please refer to the supplementary materials for the TCN architecture and training details.

\begin{figure}[t!]
    \centering
    \includegraphics[width=\textwidth]{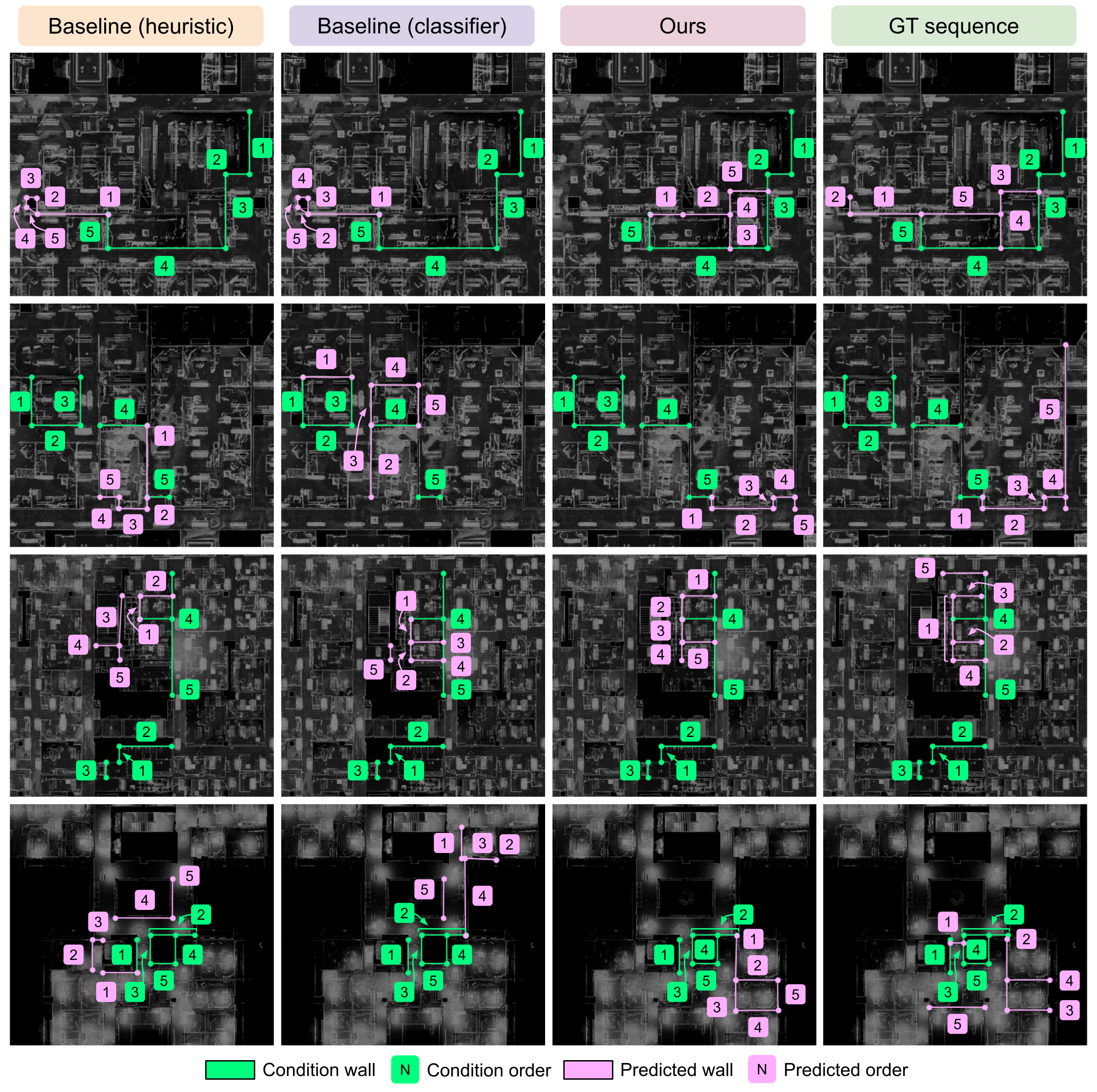}
    \vspace*{.05cm}
    \caption{Qualitative comparison of predicted sequences. We note that for the right-most column, the pink walls and ordering are GT as well.}
    \label{fig:qualitative}
\end{figure}

\subsection{Quantitative evaluations}

\figref{seq_vs_fid} presents our main results, wherein we assess the sequences predicted by three competitive methods using our proposed order metric. To generate a sequence, we select one edge from the candidates as the starting point, and subsequently predict the next $N$ steps auto-regressively, up to a maximum of 10 steps. The candidate walls can be either ground truth (GT) or predicted ones. For the latter, we use raw predictions without any post-processing to maintain high recall. As evident, our method outperforms the others on both sets of walls, despite being trained solely on GT walls and sequences.

We evaluate the impact of our modifications to the deformable attention module (Sect. \ref{sec:candidate_wall}).
As shown in \tabref{attn_samples}, we vary the number of reference points and calculate precision/recall at different distance thresholds along with Intersection over Union (IoU) scores.
To compute wall width accuracy, we collect matched walls under threshold of 30 inches, and consider width prediction to be correct if it's within 3 inches.
The method of sampling only one point is equivalent to the original HEAT architecture.
The results indicate that utilizing more points generally leads to superior performance.

Lastly, we investigate the effect of history length on the next wall prediction. More accurate and confident predictions should result from providing a longer historical context. To do so, we provide GT history of different lengths, and predict from the remaining GT edges. In \tabref{entropy_acc}, we observe that as the history length increases, both the accuracy of the predicted next wall and the entropy decrease, thereby confirming our hypothesis.
%\figref{seq_vs_fid} contains our main results, where we evaluate the predicted sequences from the three competing methods using our proposed order metric. To generate a sequence, we choose one edge from candidates as the start of the sequence, and auto-regressively predict the next $N$ steps, up to 10. The candidate walls can either be GT or predicted ones, for the latter we use raw predictions without post-processing to maintain higher recall. As we can see, our method performs better on both sets of walls, despite only trained on GT walls and sequences.

%We also evaluate the effect of our proposed modification to the deformable attention module. In \tabref{attn_samples}, we vary the number of reference points during training, and compute precision/recall at different distance thresholds along with IoU scores. Sampling only one point is equivalent to the original HEAT architecture. We can see that more points offer better performance overall.

%And finally, we study the effect of history length for next wall prediction. Theoretically, given longer history as condition, we should observe more accurate and more confident predictions. In \tabref{entropy_acc}, we observe that as history length increases, the accuracy of the predicted next wall increases and the entropy decreases, confirming our hypothesis.

\subsection{Qualitative evaluations}

% The video shows a realistic modeling scenario, with our system operating in both automatic and assistive modes.

Our system is designed to be interactive. We refer to the supplementary video for a demonstration of our system. The video depicts a real-world modeling scenario in which our system is integrated within Revit and supports the user by predicting future modeling sequences. Both automatic and assistive modes are demonstrated to highlight the flexibility of our tool.

\figref{qualitative} illustrates some of the wall sequence predictions, where a ground truth (GT) sequence is a condition. Across all the columns, the green edges represent GT walls. In the first three columns, the pink edges correspond to the reconstructed results; for the rightmost column, they signify GT walls. For ease of visualization, the history and predicted lengths are capped at five. Our method yields a sequence ordering that more closely mirrors the GT sequence. In the first row, our method intuitively grasps the intention of an architect, choosing to model the smaller enclosed area first. In the second row, our sequence ordering closely matches the GT, with the only deviations occurring in the reconstruction results.

%As our system is interactive, we refer to the supplementary video for a demo of our system. The video shows a realistic modeling scenario, where our system is integrated inside Revit and assists the user by predicting future modeling sequences. We demonstrate both the automatic and assistive modes, showing flexibility of our tool.

%\figref{qualitative} visualizes some of the wall sequence predictions, where a GT sequence is provided as condition. In all of the columns, green edges indicate GT walls. In the left three columns, pink edges are the reconstructed results; the for right most column they are GT walls. The history and predicted lengths are limited to 5 for ease of visualization. We can see that our method obtains more similar ordering compared to the GT sequence. In the first row, our method understands the architect's intention and chooses to model the smaller enclosed area first. In the second row, our ordering is almost the same as the GT, except differences in reconstruction results.

%% file: tables/attn_samples.tex
\begin{table}[]
\centering
\caption{Effect of number of reference points for deformable attention}
\label{tab:attn_samples}
\vspace*{.3cm}
\resizebox{\textwidth}{!}{%
\begin{tabular}{cccccccccccccccll}
\toprule
 &
   &
  \multicolumn{3}{c}{5 inches} &
   &
  \multicolumn{3}{c}{15 inches} &
   &
  \multicolumn{3}{c}{30 inches} &
   &
   &
   &
  \multicolumn{1}{c}{\multirow{2}{*}{\begin{tabular}[c]{@{}c@{}}Width\\ Acc.\end{tabular}}} \\ \cline{3-5} \cline{7-9} \cline{11-13}
\# points &  & Prec. & Recall & F-1   &  & Prec. & Recall & F-1   &  & Prec. & Recall & F-1   &  & IoU           &  & \multicolumn{1}{c}{} \\ \cline{1-1} \cline{3-5} \cline{7-9} \cline{11-13} \cline{15-15} \cline{17-17} 
1         &  & 59.29 & 41.01  & 48.09 &  & 67.1  & 45.95  & 54.09 &  & 70.23 & 48.03  & 56.57 &  & 0.3           &  & 79.56                \\
8         &  & 59.42 & 43.51  & 49.6  &  & 66.82 & 48.41  & 55.4  &  & 69.53 & 50.45  & 57.7  &  & \textbf{0.34} &  & 78.85                \\
16 &
   &
  \textbf{59.77} &
  \textbf{43.63} &
  \textbf{49.88} &
   &
  \textbf{67.24} &
  \textbf{48.55} &
  \textbf{55.73} &
   &
  \textbf{70.42} &
  \textbf{50.9} &
  \textbf{58.41} &
   &
  \textbf{0.34} &
   &
  \textbf{80.16} \\ \bottomrule 
\end{tabular}%
}
\end{table}

%% file: tables/entropy_acc.tex
\begin{table}[]
\centering
\caption{Entropy and accuracy vs history length}
\label{tab:entropy_acc}
\vspace*{.3cm}
\resizebox{0.8\textwidth}{!}{%
\begin{tabular}{@{}cccccccccc@{}}
\toprule
Hist. len & 1     & 2     & 3     & 4     & 5     & 6     & 7     & 8     & 9     \\ \midrule
Accuracy  & 25.17 & 27.41 & 28.48 & 29.45 & 30.21 & 30.93 & 31.72 & 32.35 & 32.72 \\
Entropy   & 2.32  & 2.26  & 2.19  & 2.15  & 2.13  & 2.09  & 2.05  & 2.02  & 2.01  \\ \bottomrule
\end{tabular}%
}
\end{table}

%% file: sections/07-conclusion.tex
\section{Limitations and conclusion}
\label{sec:conclusion}

\begin{figure}[t!]
    \centering
    \includegraphics[width=\textwidth]{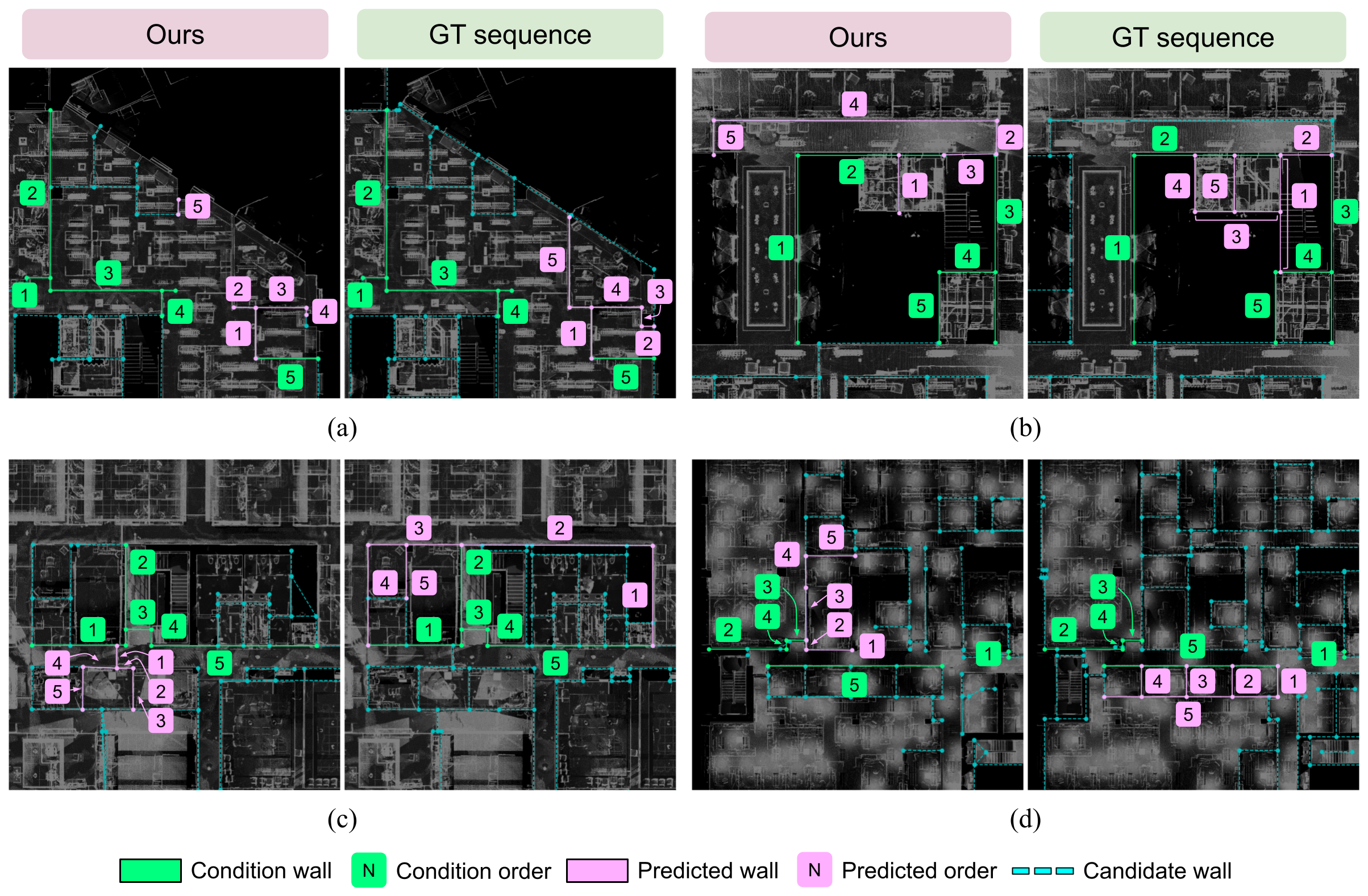}
    \vspace*{.05cm}
    \caption{Failure cases of our method. (a) and (b) suffers from poor reconstruction results; (c) and (d) demonstrates incorrect predicted order.}
    \label{fig:failure}
\end{figure}

In this paper, we have introduced a neural network architecture for the prediction of natural sequences, a Scan2BIM assistive system seamlessly integrated with professional CAD software, and an extensive Scan2BIM dataset.
%we introduce an assistive system for the Scan-to-BIM task, seamlessly integrated into professional modeling software. 
% Our method showcases strong performance in reconstruction tasks and surpasses two baseline models in next-wall prediction. With an understanding that there is room for further development, particularly in achieving production-level quality in reconstructions, we pledge to continue this advancement by making all relevant data, models, and code accessible to the research community.
Our method showcases strong performance in reconstruction tasks and surpasses two baseline models in next-wall prediction.
However, our method still exhibits failure cases with reconstructions and order predictions, which need to be seriously considered before real-world usage (see \figref{failure} for examples).
With an understanding that there is room for further development, we pledge to continue this advancement by making all relevant data, models, and code accessible to the research community.

%This paper proposes an assistive system for the task of Scan-to-BIM, which is directly integrated into the modeling software. We demonstrate that our method obtains good reconstruction performance, and is superior in the task of next wall prediction compared to two baselines. We recognize that there is still a gap in assistive modeling in the industry, as the reconstructions are not at production level. We leave such improvements as part of our future work.

% \rodg{mentioning RLHF to address some of the limitations in the current work?}

\mysubsubsectiona{Acknowledgment}
This research is partially supported by NSERC Discovery Grants with Accelerator Supplements and the DND/NSERC Discovery Grant Supplement, NSERC Alliance Grants, and the John R. Evans Leaders Fund (JELF).
This research was enabled in part by support provided by BC DRI Group and the Digital Research Alliance of Canada (\url{alliancecan.ca}).
And finally, we want to thank all the architects who helped debug our Revit plugins and participated in the data collection effort.